%% file: PaperForReview.tex
\documentclass[10pt,twocolumn,letterpaper]{article}

\usepackage{wacv}              % To produce the CAMERA-READY version
\usepackage{graphicx}
\usepackage{amsmath}
\usepackage{amssymb}
\usepackage{booktabs}
\usepackage{adjustbox}
\usepackage{float}

\usepackage{algpseudocode}
\usepackage[linesnumbered,ruled,vlined]{algorithm2e}

\usepackage[pagebackref,breaklinks,colorlinks]{hyperref}

\usepackage[capitalize]{cleveref}
\crefname{section}{Sec.}{Secs.}
\Crefname{section}{Section}{Sections}
\Crefname{table}{Table}{Tables}
\crefname{table}{Tab.}{Tabs.}

\def\wacvPaperID{1734} % *** Enter the WACV Paper ID here
\def\confName{WACV}
\def\confYear{2025}

\begin{document}

%%%%%%%%% TITLE - PLEASE UPDATE
\title{Feature Space Perturbation: A Panacea to Enhanced Transferability Estimation}

% \author{Prafful Kumar Khoba\\
% UQ–IITD Research Academy\\
% Delhi, India\\
% {\tt\small qiz228274@iitd.ac.in}
% % For a paper whose authors are all at the same institution,
% % omit the following lines up until the closing ``}''.
% % Additional authors and addresses can be added with ``\and'',
% % just like the second author.
% % To save space, use either the email address or home page, not both
% \and
% Zijian Wang\\
% The University of Queensland\\
% Brisbane, Australia\\
% {\tt\small zijian.wang@uq.edu.au}
% \and
% Chetan Arora\\
% IIT Delhi\\
% Delhi, India\\
% {\tt\small chetan@cse.iitd.ac.in}
% \and
% Mahsa Baktashmotlagh\\
% The University of Queensland\\
% Brisbane, Australia\\
% {\tt\small m.baktashmotlagh@uq.edu.au}}

\author{
Prafful Kumar Khoba$^{1}$ \quad Zijian Wang$^{2}$ \quad Chetan Arora$^{3}$ \quad Mahsa Baktashmotlagh$^{2}$ \\
$^{1}$UQ–IITD Research Academy, New Delhi, India \\
$^{2}$The University of Queensland, Brisbane, Australia \\
$^{3}$Indian Institute of Technology Delhi, New Delhi, India \\
{\tt\small qiz228274@iitd.ac.in, zijian.wang@uq.edu.au, chetan@cse.iitd.ac.in, m.baktashmotlagh@uq.edu.au}
}

\maketitle

\begin{figure*}[!h]
  \centering
  \includegraphics[width=\textwidth]{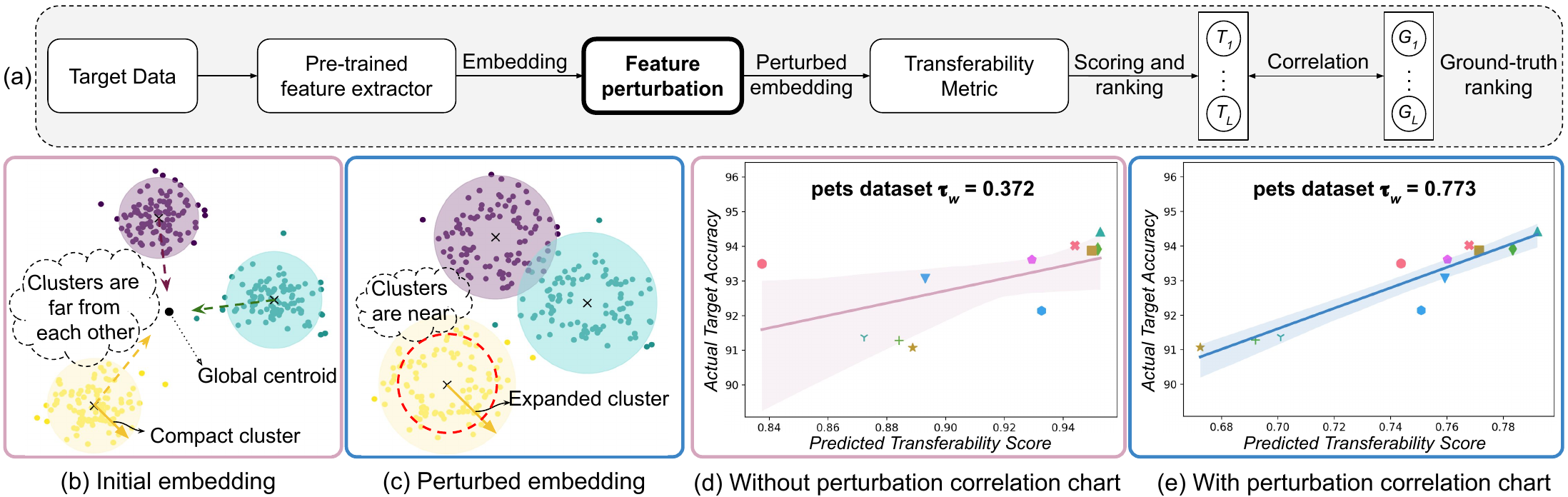}
  \caption{Illustration of our feature perturbation method for transferability estimation. (a) provides a flowchart outlining the process of enhancing transferability estimation, with bold elements representing our perturbation steps. The remaining part of the flowchart illustrates the process traditionally employed in existing transferability estimation work. (b) shows initial embeddings with significant inter-class separation and compact intra-class clustering, typical of supervised models. (c) displays embeddings after our feature perturbation.
  (d) and (e) present actual correlation charts and weighted Kendall correlation coefficient $\tau_w$ on the pets dataset. Correlation chart depicts the predicted rankings versus actual rankings before and after perturbations, where each symbol in these charts represents a model. The shift from lower to higher correlation values highlights the improved accuracy of model rankings after applying our perturbation method.}
  \label{fig:teaser}
  \vspace{-1em}
\end{figure*}

%%%%%%%%% ABSTRACT
\begin{abstract}
    Leveraging a transferability estimation metric facilitates the non-trivial challenge of selecting the optimal model for the downstream task from a pool of pre-trained models. Most existing metrics primarily focus on identifying the statistical relationship between feature embeddings and the corresponding labels within the target dataset, but overlook crucial aspect of model robustness. This oversight may limit their effectiveness in accurately ranking pre-trained models. To address this limitation, we introduce a feature perturbation method that enhances the transferability estimation process by systematically altering the feature space. Our method includes a Spread operation that increases intra-class variability, adding complexity within classes, and an Attract operation that minimizes the distances between different classes, thereby blurring the class boundaries.
   Through extensive experimentation, we demonstrate the efficacy of our feature perturbation method in providing a more precise and robust estimation of model transferability. Notably, the existing LogMe method exhibited a significant improvement, showing a 28.84$\%$ increase in performance after applying our feature perturbation method. The implementation is available at \url{https://github.com/prafful-kumar/enhancing_TE.git}
\end{abstract}

\section{Introduction}

Transfer learning enables the application of knowledge acquired from one task to enhance performance on another with only minimal additional training, typically through fine-tuning \cite{zhuang2020comprehensive}. At the core of this approach is the challenge of transferability estimation, which aims to predict how well the pre-trained models adapt when applied to new, target datasets without fine-tuning. However, the main challenge lies in determining the most appropriate model for a specific task without individually fine-tuning each candidate model, which can be a resource-intensive and time-consuming task, especially for large target datasets.

To address this issue, researchers have proposed various metrics \cite{nguyen2020leep, tran2019transferability, li2021ranking,you2021logme,shao2022not,DBLP:conf/iccv/WangLZHB23,Pandy2022TransferabilityEstimation, zhang2024model} based on the distinctive properties of the pre-trained model, known as the transferability estimation metric. The core idea behind many of these metrics is to establish a statistical relationship between the feature embedding of pre-trained models and the labels of the samples, thereby generating a transferability score for each model. Based on the score, these metrics aim to predict the actual rank of these models on the target dataset, with the goal of achieving a high correlation between the predicted rank and the actual rank, as shown in Fig. \ref{fig:teaser}(a). However, while these metrics effectively measure adaptability, they often overlook how models handle disruptions in the structure of the embeddings, which is crucial for assessing their robustness. To fill this gap, we propose a novel approach that perturbs both the intra-class (through the spread operation) and inter-class (through the attract operation) structures of the embeddings.
By introducing this targeted perturbation, we aim to provide a more comprehensive and realistic assessment of a model's resilience and adaptability. This method not only improves the accuracy of transferability estimations but also ensures that the selected models are genuinely capable of performing reliably in diverse and dynamic environments.

To illustrate how robustness to perturbations impacts transferability estimation, consider a toy example depicted in Fig. \ref{fig:teaser}.
The figure visually illustrates how embeddings respond to perturbations and its impact on model transferability. Initially, the embeddings display clear inter-class separation and compact intra-class configurations Fig. \ref{fig:teaser}(a), typical of supervised models' structured embeddings. After our feature perturbation method is applied Fig. \ref{fig:teaser}(b), these embeddings show expanded intra-class distributions and reduced inter-class separations, highlighting how perturbations disrupt the standard embedding structure. As traditional metrics rely adaptability of the features, they tend to produce lower transferability scores post-perturbation for all the models. Yet, the degree of score reduction varies significantly among the models. This is evident on the x-axis of the correlation chart for the pets dataset in Fig. \ref{fig:teaser}(d,e), as referenced from the results section \ref{sect:result}. For instance, if one model’s score drops drastically compared to others, it indicates a lower robustness to perturbations. This substantial decrease suggests that the model’s embeddings are overly sensitive to changes, potentially undermining its performance when applied to new datasets.
This leads to a more accurate relative ranking of these models. %These improvements are detailed in the experimental section \ref{sec:exp}, highlighting how perturbations refine the correlation between predicted and actual model rankings.

While most of the prior works focus on estimating the transferability of models under the vanilla fine-tuning schema, we argue that it may not be sufficient to address a broader use case. Recent literature~\cite{kirichenko2022dfr, kumar2022fine} highlights that while vanilla fine-tuning can achieve higher in-distribution (ID) test accuracy, it sacrifices the out-of-distribution (OOD) robustness compared with alternative fine-tuning strategy (\textit{i.e.}, linear probing). The trade-off between ID and OOD test accuracy urges transferability estimation metrics to consider a wider range of fine-tuning schemes. Therefore, in this work, we assess the transferability estimation metric for three distinct fine-tuning strategies: \textit{vanilla fine-tuning}, which updates all model parameters; \textit{last block fine-tuning (LBFT)}, which updates only the parameters of the last block and final linear layer; and \textit{linear fine-tuning (LFT)}, which focuses on updating the last fully connected layer. In summary, our contribution is threefold:

% Note that we avoid applying our perturbation technique to self-supervised models. This is because self-supervised models do not rely on labeled data to learn representations \cite{sela, simclr-v1, simclr-v2, byol, infomin}. Whereas our feature perturbation method, which relies on class label information, may disrupt the meaningful structures they learn by erasing valuable geometric configurations of the embeddings. Unlike supervised models, which follow a discriminative pattern, self-supervised models do not, as shown in Fig. \ref{fig:ss_vs_s}. This complexity makes it difficult to create an effective perturbation strategy for self-supervised models. As a result, our research is dedicated to improving transferability estimation for supervised models only. However, we have developed LDA-based metrics for self-supervised models that surpass many of the previous transferability estimation metrics. In summary, our contribution is threefold:

\begin{itemize}
    \item  
    Our feature perturbation method effectively perturbs the feature embedding spaces of pre-trained supervised models, significantly improving the rank estimation of these pre-trained models by transferability estimation metric. This approach led to a notable 28.84$\%$ performance boost in the LogMe \cite{you2021logme} method.
    \item 
    Our findings highlight that existing metrics are primarily effective for vanilla fine-tuning, pointing out the necessity for more adaptable solutions for diverse fine-tuning techniques. Our feature perturbation method significantly enhances transferability estimation across key fine-tuning strategies, specifically vanilla and LBFT. Its effectiveness across these strategies and compatibility with all current transferability estimation metrics underscores its broad applicability and versatility. 
    \item 
   Unlike supervised models, self-supervised models do not rely on labeled data and are particularly sensitive to disruptions in their geometric embedding structures. Therefore, we avoid applying our class-based perturbation strategy to these models. Instead, we have developed a Linear Discriminant Analysis \cite{balakrishnama1998linear} (LDA)-based metric, specifically tailored for self-supervised models, that significantly outperforms traditional baselines across all fine-tuning variants.

\end{itemize}
% Our analysis demonstrates the superior performance of simpler architectures, like Linear Discriminant Analysis (LDA) \cite{balakrishnama1998linear}, over state-of-the-art methods for self-supervised models across all fine-tuning variants. 

\section{Related Work}

% \subsection{Transferability estimation techniques}

Prior work can be categories into label-based method and source-embedding based method. Log Expected Empirical Predictor (LEEP) \cite{nguyen2020leep} and Negative Conditional Entropy (NCE) \cite{tran2019transferability} are two transferability estimation metrics that rely on information from both the source and target label sets. % One drawback of label-comparison based methods is their sensitivity to changes in the model's head (the last layer), which can affect the output scores of these metrics. Instead of using the head of the pre-trained model, source-embedding based approaches tend to use the feature extractor for creating transferability estimation metric.
Passing target data through a pre-trained model's feature extractor yields target embeddings for source-embedding methods. These methods include $\mathcal{N}$LEEP \cite{li2021ranking}, LogMe \cite{you2021logme}, SFDA \cite{shao2022not}, NCTI \cite{DBLP:conf/iccv/WangLZHB23}, and GBC \cite{Pandy2022TransferabilityEstimation}.

$\mathcal{N}$LEEP \cite{li2021ranking} is build to remove the drawbacks of LEEP, i.e., it does not uses source head. It utilizes Principal Component Analysis (PCA) \cite{wold1987principal} to reduce the dimensionality of the data, followed by fitting a Gaussian Mixture Model (GMM) \cite{reynolds2009gaussian} to the target data embedding. LogMe \cite{you2021logme} assesses model transferability by modeling each target label as a linear model with Gaussian noise using maximum evidence to evaluate how well pre-trained model features fit target labels. GBC \cite{Pandy2022TransferabilityEstimation} metric evaluates how much classes in the target data's embedding space overlap; a higher GBC value indicates greater overlap. SFDA, presented in \cite{shao2022not}, is a transferability metric which employs Fisher Discriminant Analysis (FDA) \cite{mika1999fisher} and introduces ConfMix. FDA works by finding a linear data transformation that maximizes the separation between classes using within-class scatter and between-class scatter followed by applying ConfMix, a mechanism to generate challenging negative samples. NCTI \cite{DBLP:conf/iccv/WangLZHB23} metric quantifies the gap between a model's embedding and the ideal neural collapse embedding. 

% A notable attempt by Li et al. \cite{li2023exploring} introduces the Potential Energy Decline (PED) approach, a physics-inspired method aimed at better capturing the dynamics of model adaptability by modeling the interactive forces that influence the fine-tuning process. \cite{menta2024active} enhances transferability estimation by selecting the most informative subset from the target dataset using the pre-trained model's internal and output representations. However, none of the existing works adequately address the aspect of robustness in transferability estimation. Our study introduces a vital perspective, crucial for developing more reliable transferability estimation metrics that effectively incorporate both adaptability and robustness.

Li et al. \cite{li2023exploring} proposed Potential Energy Decline (PED), a physics-inspired method modeling interactive forces to capture model adaptability during fine-tuning. Menta et al. \cite{menta2024active} improved transferability estimation by selecting informative subsets of the target dataset using pre-trained model representations. However, existing methods overlook robustness in transferability estimation. Our study addresses this gap by introducing a framework that integrates adaptability and robustness for more reliable transferability metrics.
\section{Methodology}

In this section, we outline the problem definition, our proposed method, transferability estimation metrics, and the evaluation criteria for model selection. For the sake of simplicity, we use image classification as our primary task throughout the paper.

\subsection{Preliminaries}

\textbf{Problem definition:}
Let $\mathcal{T} = \{X,Y\}$ represent the target dataset, and $\{\phi_l\}_{l=1}^L$ denote $L$ the pre-trained feature-extractors, with $l$ being the index. Using transferability estimation, given by $\mathcal{M}$, our goal is to rank these pre-trained models based on their performance on a given target dataset. We assess the transferability of each pre-trained model using a specific metric $\mathcal{M}$ that produces a numerical score, denoted as $T_l$. Essentially, a higher score for $T_l$ implies that the model $\phi_l$ is more likely to perform effectively on the given target dataset. 

\noindent\textbf{Ground truth:}
To establish a reliable basis for comparing pre-trained models, a process of fine-tuning each model on the target dataset is conducted, accompanied by a thorough exploration of different hyperparameters (i.e., learning rate and weight decay). Test accuracy obtained from this fine-tuning is then utilized as a `ground truth' for the ranking of these models. The fine-tuning performance of each model is notated as $\{G_l\}_{l=1}^L$, serving as a benchmark for evaluating the model ranking by transferability estimation metric.

\subsection{Proposed Approach}
\label{sec:SA}

Source-embedding based transferability estimation metrics\cite{shao2022not, DBLP:conf/iccv/WangLZHB23,li2021ranking,Pandy2022TransferabilityEstimation,you2021logme} leverage features extracted from pre-trained models to estimate their adaptability to a downstream task. Our proposed feature perturbation method systematically tests the robustness of model embeddings by applying controlled perturbations. Recognizing that the embedding structure varies with the feature extractor and the target dataset, our method avoids a one-size-fits-all perturbation. Instead, it dynamically adjusts the perturbation magnitude to preserve the meaningful structure of embeddings. We achieve this through two operations called spread and attract (SA), which are described below:

\noindent\textbf{Spread operation:} One of the desirable properties of features in supervised learning is high intra-class compactness \cite{bengio2013representation}. The Spread operation deliberately perturbs the internal structure of each class by increasing the intra-class variance, effectively decreasing intra-class compactness. This perturbation pushes examples that were previously near the centroid—and thus easier to classify—further away, increasing their variability and making them more challenging to classify correctly. Specifically, for every class, we compute the centroid of the class \( \mathbf{C}_u\) as given by Eq. \ref{eq:centroid} and displace each data point uniformly away from its centroid, given by:

\begin{equation}
\mathbf{\hat{X}}_{\text{spread}, u} = \mathbf{\hat{X}}_u + \left( \frac{\mathbf{\hat{X}}_u - \mathbf{C}_u}{\|\mathbf{\hat{X}}_u - \mathbf{C}_u\|_2} \right)
\label{eq:spread}
\end{equation}

\begin{equation}
\mathbf{C}_u = \frac{1}{n_u} \sum_{i=1}^{n_u} \mathbf{\hat{X}}_{u,i}
\label{eq:centroid}
\end{equation}

where \(\mathbf{\hat{X}}_{\text{spread}, u}\) represents the feature embeddings for class \(u\) after the spread operation, \(\mathbf{C}_u\) is the centroid of class \(u\), \(\mathbf{\hat{X}}_{u}\) is the reduced-dimensionality embedding of class \(u\) obtained after applying PCA on the extracted feature embeddings, and \( n_u \) is the number of samples in class \( u \).

\noindent\textbf{Attract operation:}  Another desirable property of features in supervised learning is high inter-class separability\cite{bengio2013representation}. Supervised models for classification tasks generally learn clear boundaries between classes in feature space. The attract operation aims to diminish this separation by adjusting class embeddings according to the distances between their centroids and variance, thereby perturbing the feature space to create closer inter-class proximities. This operation tests the resilience of the model's embeddings in scenarios where class boundaries are intentionally blurred.
The attract operation is formulated as:

\begin{equation}
\mathbf{\hat{X}}_{\text{attract}_{u}} = \mathbf{\hat{X}}_{\text{spread}_{u}} + \alpha \cdot \mathbf{Disp}_{uv}
\label{eq:attract}
\end{equation}

where, \(\mathbf{\hat{X}}_{\text{attract}_{u}}\) is the feature embedding of the class \textit{u} after applying the attract method to \(\mathbf{\hat{X}}_{\text{spread}_{u}}\), 
$\alpha$ is a hyper-parameter that modulates the magnitude of the displacement towards the other class centroids, and $\mathbf{Disp}_{uv}$ is: 

\begin{equation}
\mathbf{Disp}_{uv} = \sum_{v \neq u} \left( \frac{\mathbf{D}_{uv}}{\|\mathbf{D}_{uv}\|_2} \right) \cdot \left( \|\mathbf{D}_{uv}\|_2 - (\sigma \cdot R_u + \sigma \cdot R_v) \right) \\ \nonumber
\label{eq:displacement}
\end{equation} 

\begin{equation}
R_u = \sqrt{\frac{1}{n_u} \sum_{i=1}^{n_u} \| \mathbf{X}_{u,i} - \mathbf{C}_u \|^2}
\label{eq:standard_deviation}
\end{equation}

Here, $\mathbf{D}_{uv} = \mathbf{C}_u - \mathbf{C}_v$ is the distance vector between the centroids \(\mathbf{C}\) of two different classes, \(u\) and \(v\). $\sigma$ is a hyper-parameter that adjusts the sensitivity of the displacement to the variability within class embeddings, as measured by $R_u$ and $R_v$. 
Attract operation ensures that the perturbations are proportionate to the natural variability within and between the classes. This approach helps to preserve the meaningful structure of the embeddings while testing their resilience against perturbations.

\begin{algorithm} [!h]
\caption{SA algorithm}
\label{alg:SA}
\KwIn{Target dataset $\mathcal{T}=\{X,Y\}$, pre-trained models $\{\phi_l\}_{l=1}^L$, hyper-parameters $\sigma$, $\alpha$, transferability metric $\mathcal{M}$}
\KwOut{Transferability scores $T_l$ for each model}
\For{$l=1$ \textbf{to} $L$}{
    $\hat{X} \leftarrow$ PCA($\phi_l(X)$) \\
    \For{each class $k$ in $Y$}{
        $C_k \leftarrow$ Compute centroid of $\hat{X}_k$ \\
        \For{each point $X_{k,i}$ in class $k$}{
            $D_{k,i} \leftarrow \frac{X_{k,i} - C_k}{\|X_{k,i} - C_k\|_2}$ \\
            $X_{\text{spread}_{k,i}} \leftarrow X_{k,i} + D_{k,i}$ \tcp{Spread}
        }
    }
    \For{each class pair $(u, v)$}{
        $D_{uv} \leftarrow C_u - C_v$ \\
        $R_u \leftarrow \| \text{Std}(\hat{X}_u) \|_2$, $R_v \leftarrow \| \text{Std}(\hat{X}_v) \|_2$ \\
        \For{each $X_{\text{spread}_{u,i}}$ in class $u$}{
            $Disp_{uv,i} \leftarrow \sum_{v \neq u} \frac{D_{uv}}{\|D_{uv}\|_2} \cdot (\|D_{uv}\|_2 - (\sigma \cdot R_u + \sigma \cdot R_v))$ \\
            $X_{\text{attract}_{u,i}} \leftarrow X_{\text{spread}_{u,i}} + \alpha \cdot Disp_{uv,i}$ \tcp{Attract}
        }
    }
    $T_l \leftarrow \mathcal{M}(X_{\text{attract}}, Y)$
}
Rank models based on $T_l$.
\end{algorithm}

\begin{figure}[!h]
  \centering
  \includegraphics[width=0.90\columnwidth]{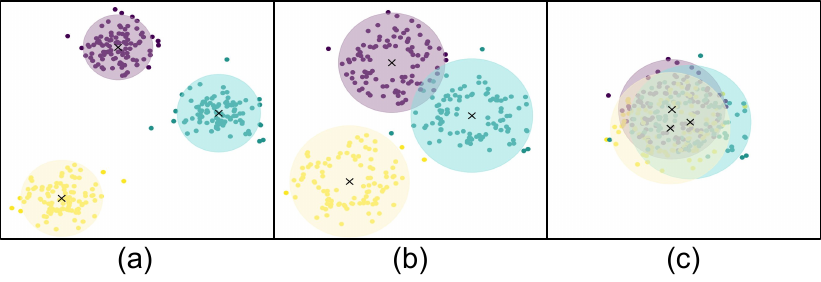}
  \vspace{-2ex}
  \caption{Demonstrating the importance of controlled perturbation in feature space manipulation, using a toy example. (a) Represents the initial target embedding. (b) depict an appropriate amount of feature perturbation, while (c) demonstrate excessive levels of feature perturbation.}
  \label{fig:hyp}
  \vspace{-1ex}
\end{figure}

 \noindent\textbf{The importance of controlled perturbation:}
Fig. \ref{fig:hyp}(b) demonstrates an optimal level of perturbation that preserves the necessary balance and structural integrity required for accurate transferability estimation. Conversely, Fig. \ref{fig:hyp}(c) shows how suboptimal perturbation settings can result in excessive alterations, complicating the feature space and hindering transferability assessment. These examples underline the crucial role of carefully calibrated perturbations in ensuring reliable model evaluations.

\subsection{Overall Objective}

Our feature perturbation method systematically alters the feature embedding representation for target dataset $\mathcal{T}=\{X,Y\}$ using $\{\phi_l\}_{l=1}^L$ feature extractor of the pre-trained model. We then compute the transferability score for each model $\{T_l\}_{l=1}^L$ using the metric $\mathcal{M}$, which assesses the perturbed embeddings $\mathbf{\hat{X}}_{\text{attract}}$ against the target labels $\mathbf{Y}$, as shown in the equation below. 
\begin{equation}
\label{eq:transferability_estimation}
    T_l = \mathcal{M}(\mathbf{\hat{X}}_{\text{attract}}, \mathbf{Y})
\end{equation}
For each feature extractor $\phi_l$, the algorithm extracts feature embeddings and applies PCA to get a reduced-dimensionality representation, $\hat{X}$. This step is crucial for managing computational complexity and focusing on the most informative features of the embeddings. The algorithm for our perturbation method is given in Algorithm \ref{alg:SA}. 
\begin{table*}[t!]\setlength\tabcolsep{5pt}
\footnotesize
\centering
\caption{Performance comparison (average weighted Kendall $\tau_w$) between original and enhanced frameworks for vanilla fine-tuning on supervised models. For a pair of rows, the original metric is presented first followed by the corresponding enhanced metric. Best results are highlighted in bold. Our framework consistently outperforms the original framework across all metrics.} 
\label{tab:sup_fine-tune}
\begin{adjustbox}{max width=\textwidth}
\begin{tabular}{l|ccccccccccc|c}
    \toprule
    Method          & Aircraft & Caltech-101 & Cars  & CIFAR10 & CIFAR100 & DTD   & Flowers & Food-101 & Pets  & Sun & VOC & Average \\ 
    \midrule
    $\mathcal{N}$LEEP\cite{li2021ranking}           & -0.449   & \textbf{0.769}       & 0.602 & 0.783   & 0.717    & 0.796 & 0.295   & 0.581    & 0.511 & \textbf{0.944} & 0.710 & 0.569 \\
    \textbf{SA} + $\mathcal{N}$LEEP& \textbf{0.236}   & 0.626       & \textbf{0.763} & \textbf{0.910}    & \textbf{0.843}    & \textbf{0.836} & \textbf{0.435}   & \textbf{0.657}   & \textbf{0.829} & 0.790 & \textbf{0.828}   & \textbf{0.704} \\
    \hline
    LogME\cite{you2021logme}           & 0.439    & 0.497       & \textbf{0.605} & 0.852   & 0.725    & 0.700 & 0.147   & 0.385    & 0.411 & 0.511 & 0.695 & 0.542 \\
    \textbf{SA} + LogME & \textbf{0.442}    & \textbf{0.655}       & 0.603 & \textbf{0.924}  & \textbf{0.855}    & \textbf{0.784} & \textbf{0.743}   & \textbf{0.665}    & \textbf{0.447} & \textbf{0.788} & \textbf{0.782} & \textbf{0.698} \\
    \hline
    GBC\cite{Pandy2022TransferabilityEstimation}             & \textbf{0.423}    & 0.213       & \textbf{0.617} & 0.735   & 0.664    & 0.703 & 0.214   & 0.548    & 0.514 & 0.271 & 0.743 & 0.513 \\
    \textbf{SA} + GBC   & -0.110    & \textbf{0.473}       & 0.591 & \textbf{0.928}   & \textbf{0.789}    & \textbf{0.713} & \textbf{0.510}  & \textbf{0.711}   & \textbf{0.651} & \textbf{0.803} & \textbf{0.787} & \textbf{0.622} \\
    \hline
    SFDA\cite{shao2022not}            & \textbf{0.614}    & \textbf{0.615}       & 0.574 & \textbf{0.949}   & 0.866    & 0.575 & 0.492   & \textbf{0.815}    & 0.545 & 0.558 & 0.671 & 0.661 \\
    \textbf{SA} + SFDA  & 0.414    & 0.598       & \textbf{0.801} & 0.901   & \textbf{0.908}    & \textbf{0.816} & \textbf{0.736}   & 0.681    & \textbf{0.865} & \textbf{0.790} & \textbf{0.816} & \textbf{0.756} \\
    \hline
    NCTI\cite{DBLP:conf/iccv/WangLZHB23}            & 0.496     & \textbf{0.492}      & 0.662 & 0.843   & \textbf{0.879}    & 0.616 & 0.541   & \textbf{0.773}    & \textbf{0.867} & 0.756 & 0.741 & 0.697 \\
    \textbf{SA} + NCTI  & \textbf{0.872}   & 0.483       & \textbf{0.805} & \textbf{0.843}   & 0.878    & \textbf{0.776} & \textbf{0.714}   & 0.619    & 0.856 & \textbf{0.790} & \textbf{0.800}   & \textbf{0.767} \\
    \bottomrule
\end{tabular}
\end{adjustbox}
\end{table*}

% LBFT supervised
\begin{table*}[h]\setlength\tabcolsep{5pt}
\footnotesize
\centering
\caption{Performance comparison (average weighted Kendall $\tau_w$) between original and enhanced frameworks for LBFT on supervised models. For a pair of rows, the original metric is presented first followed by the corresponding enhanced metric. The optimal outcomes are emphasized in bold. Our framework consistently exceeds the performance of the original framework across all metrics.} 
\label{tab:sup_conv-ft}
\begin{adjustbox}{max width=\textwidth}
\begin{tabular}{l|ccccccccccc|c}
    \toprule
    Method & Aircraft & Caltech-101 & Cars & CIFAR10 & CIFAR100 & DTD & Flowers & Food-101 & Pets & Sun & VOC & Average \\ \midrule
    $\mathcal{N}$LEEP\cite{li2021ranking} & -0.415 & 0.370 & 0.159 & 0.732 & 0.803 & 0.593 & -0.035 & 0.667 & 0.505 & \textbf{0.691} & 0.512 & 0.417 \\
    \textbf{SA} + $\mathcal{N}$LEEP & \textbf{0.305} & \textbf{0.735} & \textbf{0.848} & \textbf{0.757} & \textbf{0.853} & \textbf{0.629} & \textbf{0.020} & \textbf{0.742} & \textbf{0.529} & 0.641 & \textbf{0.777} & \textbf{0.621} \\ 
    \hline
    LogME\cite{you2021logme} & \textbf{0.386} & \textbf{0.577} & 0.453 & \textbf{0.789} & 0.640 & \textbf{0.715} & 0.286 & 0.690 & 0.192 & \textbf{0.627} & 0.222 & 0.507 \\
    \textbf{SA} + LogME & 0.223 & 0.260 & \textbf{0.633} & 0.739 & \textbf{0.831} & 0.548 & \textbf{0.458} & \textbf{0.783} & \textbf{0.233} & 0.622 & \textbf{0.707} & \textbf{0.548} \\
    \hline
     GBC\cite{Pandy2022TransferabilityEstimation} & \textbf{0.676} & 0.076 & 0.476 & 0.631 & 0.751 & 0.612 &\textbf{ 0.176} & \textbf{0.790} & \textbf{0.349} & 0.395 & 0.210 & 0.467 \\
    \textbf{SA} + GBC & -0.164 & \textbf{0.541} & \textbf{0.626} & \textbf{0.656} & \textbf{0.767} & \textbf{0.650} & 0.133 & 0.749 & 0.149 & \textbf{0.641} & \textbf{0.777} & \textbf{0.502} \\
    \hline
    SFDA\cite{shao2022not} & \textbf{0.395} & 0.432 & 0.324 & 0.702 & 0.671 & 0.585 & \textbf{0.414} & 0.553 & 0.372 & 0.393 & 0.142 & 0.453 \\
    \textbf{SA} + SFDA & 0.174 & \textbf{0.754} & \textbf{0.864} & \textbf{0.765} & \textbf{0.775} & \textbf{0.768} & 0.361 & \textbf{0.708} & \textbf{0.698} & \textbf{0.610} & \textbf{0.764} & \textbf{0.658} \\
    \hline
    NCTI\cite{DBLP:conf/iccv/WangLZHB23} & 0.366 &0.441& 0.447 & 0.728 & 0.760 & 0.395 & 0.150 & 0.637 & \textbf{0.766} & \textbf{0.848} & 0.388 & 0.539 \\
    \textbf{SA} + NCTI & \textbf{0.620} & \textbf{0.652} & \textbf{0.942} & \textbf{0.739} &\textbf{ 0.789} & \textbf{0.568} & \textbf{0.415} & \textbf{0.730} & 0.685 & 0.641 & \textbf{0.792} & \textbf{0.688} \\ 
    \bottomrule
\end{tabular}
\end{adjustbox}
\end{table*}

 % linear probing supervised
\begin{table*}[h]\setlength\tabcolsep{5pt}
\footnotesize
\centering
\caption{Comparison (average weighted Kendall $\tau_w$) between original and enhanced frameworks for LFT on supervised models. In each pair of rows, the original metric is listed first, followed by the corresponding enhanced metric. The superior results are emphasized in bold.} 
\label{tab:sup_lp}
\begin{adjustbox}{max width=\textwidth}
\begin{tabular}{l|ccccccccccc|c}
    \toprule
    Method & Aircraft & Caltech-101 & Cars & CIFAR10 & CIFAR100 & DTD & Flowers & Food-101 & Pets & Sun & VOC & Average \\ \midrule
    $\mathcal{N}$LEEP\cite{li2021ranking} & 0.422 & 0.685 & \textbf{0.747} & 0.558 & 0.509 & \textbf{0.790} & \textbf{0.378} & 0.512 & \textbf{0.717} & \textbf{0.746} & 0.698 &\textbf{ 0.615} \\
    \textbf{SA} + $\mathcal{N}$LEEP & \textbf{0.615} & \textbf{0.760} & 0.328 & \textbf{0.831} & \textbf{0.705} & 0.635 & 0.285 & \textbf{0.615} & 0.677 & 0.244 & \textbf{0.824} & 0.592 \\
    \hline
    LogME\cite{you2021logme} & \textbf{0.127} & 0.247 & \textbf{0.144} & 0.490 & 0.359 & \textbf{0.721} & 0.170 & 0.229 & -0.128 & 0.198 & 0.470 & 0.275 \\
    \textbf{SA} + LogME & -0.129 & \textbf{0.363} & 0.073 & \textbf{0.907} & \textbf{0.883} & 0.634 & \textbf{0.465} & \textbf{0.625} & \textbf{0.178} & \textbf{0.262} & \textbf{0.689} & \textbf{0.450} \\
    \hline
     GBC\cite{Pandy2022TransferabilityEstimation} & \textbf{-0.048} & 0.266 & \textbf{0.124} & 0.359 & 0.373 & 0.440 & 0.121 & 0.330 & \textbf{0.209} & 0.228 & 0.531 & 0.267 \\
    \textbf{SA} + GBC & -0.308 & \textbf{0.628} & 0.044 & \textbf{0.935} & \textbf{0.818} & \textbf{0.758} & \textbf{0.304} & \textbf{0.694} & 0.096 & \textbf{0.244} & \textbf{0.824} & \textbf{0.457} \\
    \hline
    SFDA\cite{shao2022not} & 0.320 & 0.475 & \textbf{0.508} & 0.611 & 0.558 & 0.380 & \textbf{0.429} & \textbf{0.709} & 0.119 & \textbf{0.574} & 0.432 & 0.465 \\
    \textbf{SA} + SFDA & \textbf{0.585} & \textbf{0.934} & 0.381 & \textbf{0.882} & \textbf{0.905} & \textbf{0.855} & 0.307 & 0.684 & \textbf{0.751} & 0.251 & \textbf{0.812} & \textbf{0.667} \\
    \hline
    NCTI\cite{DBLP:conf/iccv/WangLZHB23} & \textbf{0.656} & 0.775 & \textbf{0.572} & 0.627 & 0.636 & 0.526 & \textbf{0.589} & \textbf{0.675} & 0.431 &\textbf{ 0.521} & 0.667 & 0.607 \\
    \textbf{SA} + NCTI & 0.182 & \textbf{0.833} & 0.381 & \textbf{0.907} & \textbf{0.840} & \textbf{0.738} & 0.423 & 0.635 & \textbf{0.706} & 0.244 & \textbf{0.839} & \textbf{0.611} \\
    \bottomrule
\end{tabular}
\end{adjustbox}
\end{table*}

\section{Experiments}
\label{sec:exp}

This section is organized into five distinct parts to evaluate the proposed feature perturbation method. Section \ref{sect:setup} outlines the experimental setup; Section \ref{sect:result} presents the results; Section \ref{sect:ablation} covers the ablation study; Section \ref{sect:hyperparameter} examines hyper-parameter sensitivity; Section \ref{sect:time} analyzes time complexity; and section \ref{sect:self-supervised} presents the result on self-supervised architectures.

\subsection{Experiment Setup}
\label{sect:setup}

\noindent\textbf{Datasets.} Our study utilizes a diverse collection of datasets commonly used in transferability estimation research \cite{shao2022not}. The collection contains fine-grained object classification dataset (\textit{i.e.}, Oxford-102 Flowers \cite{nilsback2008automated}, Food-101 \cite{bossard2014food}, Stanford Cars \cite{cars}, FGVC Aircraft \cite{maji2013fine}, Oxford-IIIT Pets \cite{parkhi2012cats}),
coarse-grained object classification dataset (\textit{i.e.}, Caltech-101 \cite{fei2004learning}, Cifar-10 \cite{cifar}, Cifar-100 \cite{cifar}, Voc2007 \cite{pascal-voc-2007} ), 
one scene classification dataset (\textit{i.e.}, Sun397 \cite{xiao2010sun}), and 
one texture classification dataset (\textit{i.e.}, DTD \cite{cimpoi2014describing}). These datasets provide a broad spectrum of challenges under various scenarios.

\subsection{Experimental Results}
\label{sect:result}

\noindent\textbf{Overview.} To evaluate the performance of transferability estimation metrics, we initiate the model pool by following the existing work~\cite{shao2022not}. Specifically, the model pool consists of 11 ImageNet pre-trained architectures, including InceptionV1 \cite{goingdeeperwithconvolutions}, InceptionV3 \cite{rethinkingtheinceptionarchitectureforcv}, ResNet50 \cite{deepresidulalearningforimagerecognition}, ResNet101 \cite{deepresidulalearningforimagerecognition}, ResNet152 \cite{deepresidulalearningforimagerecognition}, DenseNet121 \cite{denselyconnectedconvolutionalnetworks}, DenseNet169 \cite{denselyconnectedconvolutionalnetworks}, DenseNet201 \cite{denselyconnectedconvolutionalnetworks}, MobileNetV2 \cite{mobilenetv2invertedresidualsandlinearbottlenecks}, and NASNet-A Mobile \cite{mnasnetplatformawareneuralarchitecturesearchformobile}.

\noindent\textbf{Ground truth and correlation measurement.}
We follow the grid search described in \cite{shao2022not}, which selects the learning rates from $\{10^{-1}, 10^{-2}, 10^{-3}, 10^{-4}\}$ and weight decay parameters from $\{10^{-6}, 10^{-5}, 10^{-4}, 10^{-3}\}$. To ensure the robustness and reliability of our evaluation, we execute the code using five distinct seeds for each experiment and then take the average of target accuracy. For evaluation, we use weighted Kendall \cite{kendall1938new} correlation coefficient $\tau_w$ because it assigns weights to the concordant and discordant pairs based on their positions in the ranking; A high positive $\tau_w$ value indicates a strong correlation, while a negative value implies an inverse correlation.

\noindent\textbf{Performance comparison on vanilla fine-tuning.} In this part, we consider vanilla fine-tuning, which updates all parameters during the training process. We report the experimental results in Table \ref{tab:sup_fine-tune}. As can be seen from the table, SA feature perturbation method demonstrates notable improvements in transferability estimation across all previous metrics. In particular, our proposed method achieves a relative improvement of 23.86 \% on $\mathcal{N}$LEEP, 28.84\% on LogMe, 21.27\% on GBC, 14.46\% on SFDA, 10.04\% on NCTI, and with the LogMe metric benefiting the most from this approach. The table reports an average improvement of 19.69\% on previous SOTA transferability estimation methods, indicating the effectiveness of our approach.

\begin{figure}[!t]
  \centering
  \includegraphics[width=\columnwidth]{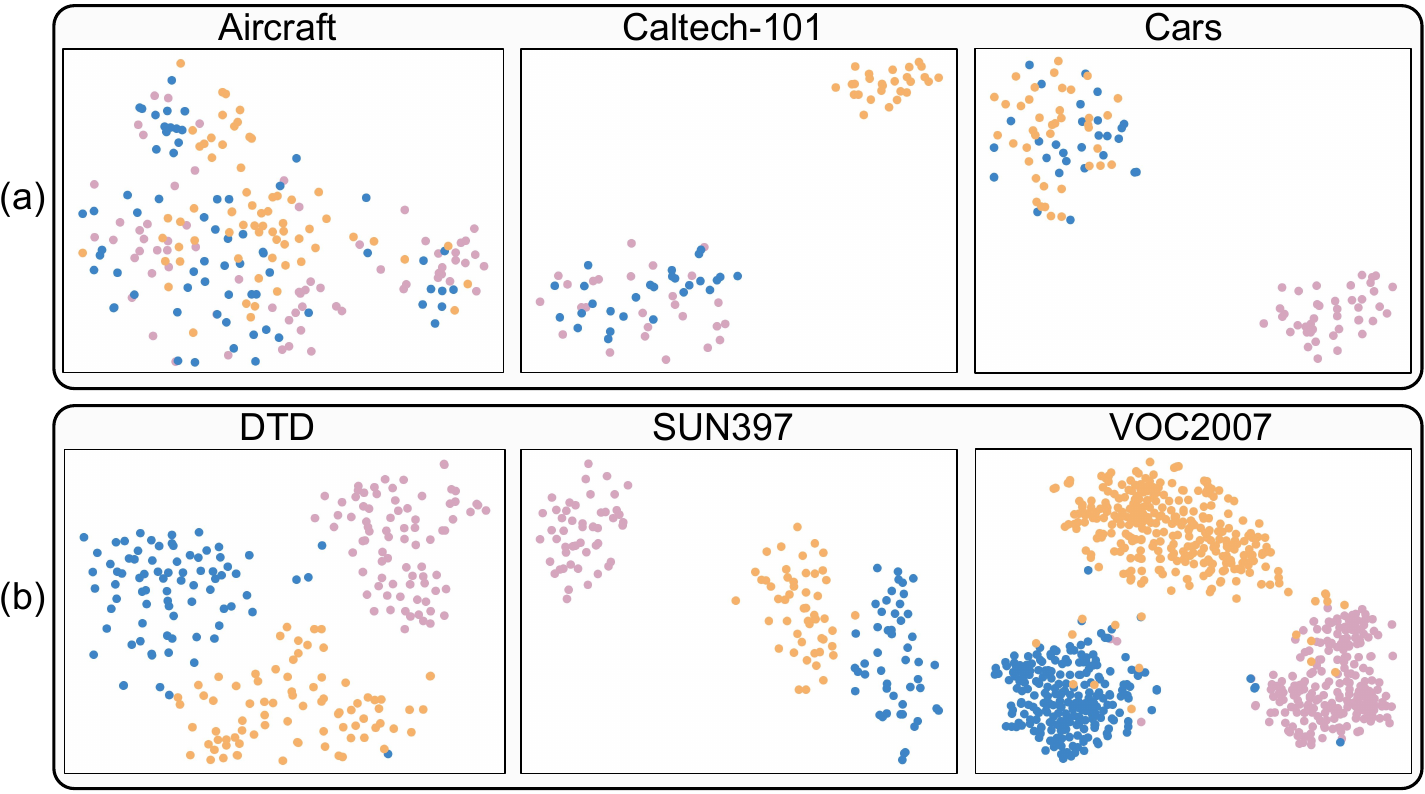}
  \caption{Visualization of ResNet50 target embeddings before feature perturbation (best viewed in color): (a) Represents datasets exhibiting mixed improvement for various metrics, shown in Table \ref{tab:sup_fine-tune}. The presence of class overlap in (a) contributes to the varied performance across metrics. In contrast, (b) depicts datasets demonstrating consistent improvement across all other evaluated metrics, facilitated by well-separated embeddings. This distinction underscores the role of embedding structure in the estimation.}
  \label{fig:result_embed}
  \vspace{-1ex}
  %\vspace{-2em}
\end{figure}

% \textcolor{red}{should we remove this para. This figure and para can develop some question in the minds of reviewers}
Meanwhile, our perturbation strategy demonstrates variably beneficial outcomes in a few datasets. For example, on Aircraft, our methods can only improve three out of five baseline metrics. We argue that embedding structure for a few datasets contains class overlap causing mixed improvements for various metrics. To understand the intrinsic difference of these datasets, we randomly sample three classes and visualize the feature distribution of (a) variably beneficial datasets and (b) consistent beneficial datasets in Fig. \ref{fig:result_embed}. One can see from the figure that the class-conditional feature distribution largely overlaps with each other. A similar pattern can be seen in other datasets like Caltech-101 and Cars. Given the well-separated feature embeddings in set (b), our perturbation method has a more pronounced effect in enhancing transferability estimation. This improvement is evident in the column corresponding to the dataset in Table \ref{tab:sup_fine-tune}. %Additional results and detailed analyses are provided in the supplementary material.

\noindent\textbf{Performance comparison on fine-tuning variants.} In this work, we study the performance of transferability estimation metrics under two alternative fine-tuning strategies, namely LBFT and LFT. LBFT updates the last convolutional block and the final linear layer of the network, and LFT only updates the final linear layer with all other parameters frozen in the training process. More details can be found in the supplementary material. 

Table \ref{tab:sup_conv-ft} shows the ranking performance of the models trained under the LBFT strategy.  Although the order of rank of the estimation metrics remains largely consistent between vanilla fine-tuning and LBFT, the overall ranking correlation drops from 0.59 to 0.47. This suggests that previous metrics could be vulnerable when applied to LBFT. Following SA feature perturbation, there is a substantial performance boost of 27.47\% for LBFT, proving the effectiveness of our feature perturbation method in enhancing accurate transferability estimation specifically for LBFT scenarios.

Compared to vanilla fine-tuning, LFT can be deployed faster, with improved performance on out-of-distribution (OOD) test samples. The ranking performance under the LFT strategy is shown in Table.~\ref{tab:sup_lp}. While a strategy like LFT does not change the structure of the feature space during training, our approach can still achieve a predominantly beneficial result. SA feature perturbation strategy improves three out of five comparison metrics, scoring 0.142 weighted Kendall $\tau_w$ increase. The results reflect that our approach is suitable to adopt in the LFT use case.

\begin{table*}[!tbh]
\setlength\tabcolsep{5pt}
\footnotesize
\centering
\caption{Performance comparison (average weighted Kendall $\tau_w$) for vanilla fine-tuning on self-supervised models.  In each column, the best results are highlighted in bold. Remarkably, LDA achieves the highest overall average weighted Kendall’s $\tau_w$ score.} 
\label{tab:ss_vanilla-ft}
\begin{adjustbox}{max width=\textwidth}
\begin{tabular}{l|ccccccccccc|c}
    \toprule
    Method & Aircraft & Caltech-101 & Cars & CIFAR10 & CIFAR100 & DTD & Flowers & Food-101 & Pets & Sun397 & VOC & Average \\ 
    \midrule
    $\mathcal{N}$LEEP\cite{li2021ranking} & -0.029 & 0.631 & 0.358 & 0.074 & 0.276 & 0.641 & 0.585 & 0.544 & \textbf{0.836} & 0.735 & -0.076 & 0.416 \\
    LogME\cite{you2021logme} & 0.223 & 0.387 & 0.387 & 0.295 & -0.028 & 0.627 & 0.718 & 0.570 & 0.704 & 0.217 & 0.121 & 0.384 \\
    GBC\cite{Pandy2022TransferabilityEstimation} & 0.070 & 0.417 & 0.464 & -0.054 & 0.237 & 0.317 & 0.701 & 0.729 & 0.484 & 0.539 & 0.161 & 0.370 \\
    SFDA\cite{shao2022not} & \textbf{0.254} & 0.526 & 0.553 & 0.619 & 0.548 & 0.815 & \textbf{0.847} & 0.685 & 0.556 & 0.732 & 0.532 & 0.606 \\
    NCTI\cite{DBLP:conf/iccv/WangLZHB23} & 0.035 & 0.643 & \textbf{0.724} & 0.546 & 0.533 & 0.715 & 0.705 & 0.892 & 0.767 & 0.697 & 0.547 & 0.619 \\
    LDA & 0.058 & \textbf{0.708} & 0.720 & \textbf{0.707} & \textbf{0.692} & \textbf{0.913} & 0.779 & \textbf{0.944} & 0.540 & \textbf{0.892} & \textbf{0.723} & \textbf{0.698} \\
    \bottomrule
\end{tabular}
\end{adjustbox}
\end{table*}

% selfsup linear probe
\begin{table*}[!tbh] 
    \setlength\tabcolsep{5pt}
    \footnotesize
    \centering
    \caption{Performance comparison (average weighted Kendall $\tau_w$) for LFT on self-supervised models. 
    The highest performing $\tau_w$ value in each column are highlighted in bold. LDA achieves the highest overall average weighted Kendall $\tau_w$ score.} 
    \label{tab:ss_lp}
    \begin{adjustbox}{max width=\textwidth}
    \begin{tabular}{l|ccccccccccc|c}
        \toprule
        Method & Aircraft & Caltech-101 & Cars & CIFAR10 & CIFAR100 & DTD & Flowers & Food-101 & Pets & Sun397 & VOC & Average \\
        \midrule
        $\mathcal{N}$LEEP\cite{li2021ranking} & 0.446 & 0.654 & 0.537 & -0.044 & 0.210 & 0.668 & 0.633 & 0.519 & 0.608 & 0.275 & 0.126 & 0.421 \\
        LogME\cite{you2021logme} & 0.569 & 0.350 & 0.629 & -0.083 & -0.250 & 0.655 & 0.653 & 0.518 & 0.487 & -0.178 & 0.037 & 0.308 \\
        GBC\cite{Pandy2022TransferabilityEstimation} & 0.498 & 0.446 & 0.717 & -0.093 & 0.147 & 0.504 & 0.702 & 0.590 & 0.422 & 0.219 & 0.320 & 0.407 \\
        SFDA\cite{shao2022not} & 0.685 & 0.582 & 0.813 & 0.275 & 0.138 & 0.717 & 0.705 & 0.693 & 0.681 & 0.426 & 0.633 & 0.577 \\
        NCTI\cite{DBLP:conf/iccv/WangLZHB23} & 0.842 & 0.661 & \textbf{0.917} & 0.275 & 0.473 & 0.699 & 0.683 & 0.846 & \textbf{0.846} & 0.308 & 0.670 & 0.656 \\
        LDA & \textbf{0.903} & \textbf{0.764} & 0.800 & \textbf{0.598} & \textbf{0.497} & \textbf{0.807} & \textbf{0.845} & \textbf{0.867} & 0.748 & \textbf{0.656} & \textbf{0.823} & \textbf{0.755} \\
        \bottomrule
    \end{tabular}
    \end{adjustbox} \vspace{-1ex}
\end{table*}

\subsection{Ablation Study}
\label{sect:ablation}

\begin{figure}
    \centering
    \includegraphics[width=0.9\columnwidth]{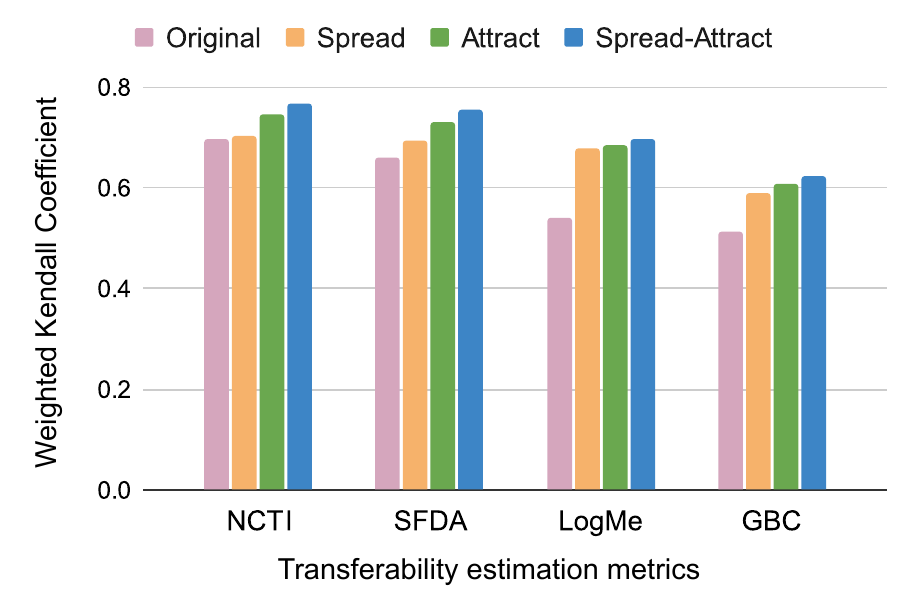}
    \vspace{-2ex}
    \caption{This figure demonstrates a bar chart that illustrates the performance improvement of various operations of feature perturbation over the original baseline. Each metric is represented by four bars, corresponding to different operations: Original, Spread, Attract, and Combined Spread-Attract, illustrating that the combined approach significantly outperform others.}
    \label{fig:ablation}
    \vspace{-3ex}
\end{figure}

To validate the effectiveness of each component in the proposed methods, we conduct an ablation study of the Spread and Attract operation on four different baseline transferability estimation metrics (\textit{i.e.,} NCTI, SFDA, LogMe, and GBC). We show the experimental results in Fig. \ref{fig:ablation}. From the figure, we can observe that the application of both the Spread and Attract operations brings improvement in ranking correlations. On average, the Spread operation yields a 10.57\% improvement, while the Attract operation provides a 14.57\% improvement. The Attract operation's superior performance can be attributed to its sophisticated approach that accounts for both the intra-class coherence and the inter-class separations before perturbation. Specifically, as defined in Eq \ref{eq:attract}, the Attract operation applies changes to the embeddings after considering the distance between class centroids (\(\mathbf{D}_{uv}\)) and the L2 norm of standard deviations within the classes (\(R_u\) and \(R_v\)). This allows for perturbations that are informed by a holistic view of the entire feature space, maintaining a delicate balance between disrupting and preserving the structural integrity essential for accurate class differentiation.
Moreover, the most substantial enhancement is observed when both operations are applied sequentially: first applying the Spread operation followed by the Attract operation. This sequential application leads to an average improvement of 17.82\% over the originally obtained weighted Kendall coefficient. This demonstrates that the synergistic effect of applying both operations sequentially is significantly greater than the impact of each operation when applied independently.

\begin{figure}
    \centering
    \includegraphics[width=\columnwidth]{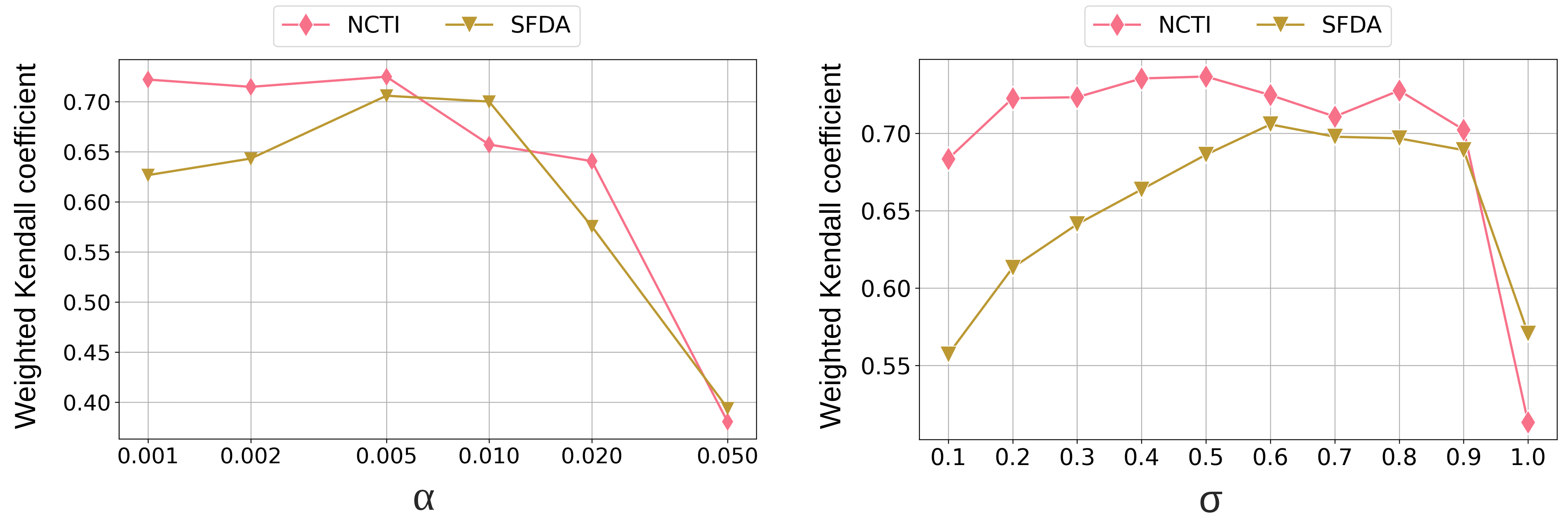}
    \vspace{-4ex}
    \caption{Hyper-parameter sensitivity analysis: The left figure showcases consistent performance ($\tau_w$) across a wide range of $\alpha$ values, at optimum $\sigma$. This observation indicates an insensitivity to hyper-parameter changes. On the other hand, the right figure illustrates limited variance in performance ($\tau_w$) across a broad spectrum of $\sigma$ values, ranging from 0.5 to 0.9, at optimum $\alpha$. %The minimal performance fluctuation observed across diverse parameter ranges underscores the algorithm's resilience, thereby emphasizing its robustness and reliability.
    } 
    \label{fig:sensitivity}
    \vspace{-3ex}
\end{figure}

\subsection{Hyper-parameter Sensitivity Analysis}
\label{sect:hyperparameter}

In this section, we assess the impact of key hyperparameters, namely $\sigma$ and $\alpha$, which govern the magnitude of perturbation within our method. The sensitivity analysis is conducted on two best-performed transferability estimation metrics (\textit{i.e.,} SFDA and NCTI) after applying the proposed SA feature perturbation technique. Fig. \ref{fig:sensitivity} shows the ranking correlation under different $\sigma$ and $\alpha$. The minimal variations in performance across a wide range of hyper-parameters highlight the algorithm's resilience, showcasing its robustness and reliability. We note that we fix the value of one hyper-parameter to tune the other one. From the figure, we can see that a similar performance trend is presented for both metrics. Specifically, the optimal performance is achieved at $\alpha$ = 0.005 and $\sigma$ = 0.6 for metrics. The analysis results demonstrate that the recommended hyper-parameters are versatile.

\subsection{Time Complexity}
\label{sect:time}

\begin{figure}
    \centering
    \includegraphics[width=0.9\columnwidth]{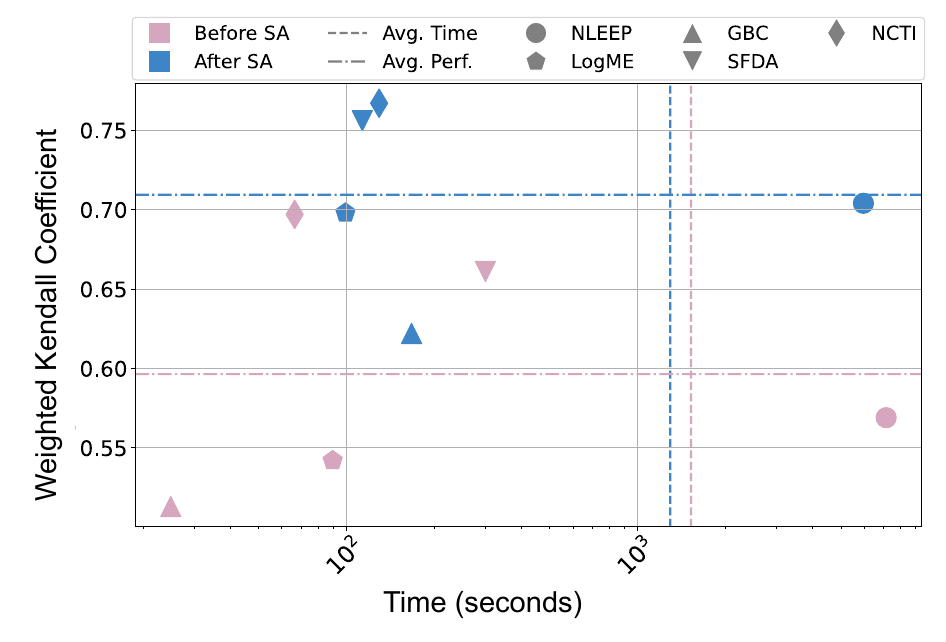}
        \vspace{-2ex}
    \caption{This figure compare the time complexity before and after applying feature perturbation techniques to vanilla fine-tuning.} 
    \label{fig:time}
    \vspace{-1ex}
\end{figure}

Our feature perturbation strategy improves ranking correlation while maintaining the same level of time complexity, as shown in Fig. \ref{fig:time}. The impact of our feature perturbation method on the time complexity of various transferability estimation metrics has yielded mixed outcomes. While some metrics have experienced a reduction in time complexity, others have seen an increase, reflecting the effect of our feature perturbation method on different evaluation approaches.
Metrics that rely heavily on the dimensionality of features such as $\mathcal{N}$LEEP, benefit from our method's dimensionality reduction, which reduces processing time. However, metrics that are less dependent on feature dimensionality do not experience the same reductions in time complexity. In fact, the introduction of SA perturbation adds computational steps, slightly increasing overall time requirements.

\subsection{Performance on Self-supervised Models}
\label{sect:self-supervised}

To evaluate the transferability estimation on the self-supervised models, we construct the pool with ResNet50 \cite{deepresidulalearningforimagerecognition} pretrained on various self-supervised methods, spanning BYOL \cite{byol}, Infomin \cite{infomin}, PCL-v1 \cite{pcl}, PCL-v2 \cite{pcl}, Sela-v2 \cite{sela}, InsDis \cite{indis}, SimCLR-v1 \cite{simclr-v1}, SimCLR-v2 \cite{simclr-v2}, MoCo-v1 \cite{moco-v1}, MoCo-v2 \cite{moco-v2}, DeepCluster-v2 \cite{deepcluster}, and SWAV \cite{swav}. In self-supervised tasks, we develop a baseline method, which leverages LDA to predict class probabilities for each sample and accumulate the probability of the correct class corresponding to samples as the transferability score. Further details can be found in supplementary material.

\begin{figure}[t!]
  \centering
  \includegraphics[width=\columnwidth]{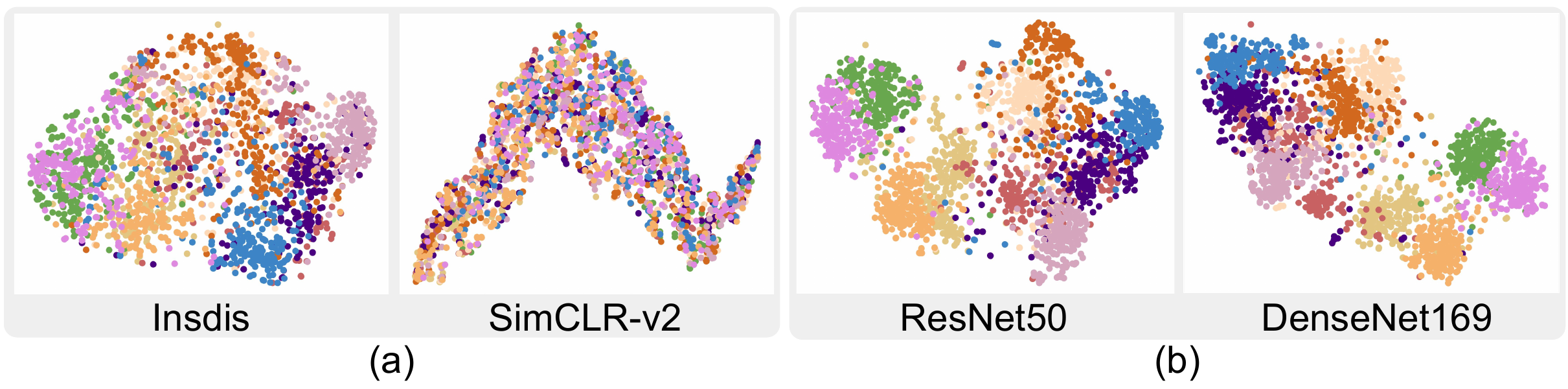}
  \caption{Comparison of CIFAR-10 embedding structures: (a) illustrates the embedding structure derived from a self-supervised model, while (b) depicts the embedding structure from a supervised model. Self-supervised models, learning without explicit label guidance, tend to capture more abstract relationships in the data, leading to embeddings with diverse patterns. In contrast, supervised models emphasize class separation, leading to a similar embedding structure for different models.}
  \label{fig:ss_vs_s}
  \vspace{-2ex}
\end{figure}

LDA-based metric achieves impressive results on transferability estimation of self-supervised models as shown in Table. \ref{tab:ss_vanilla-ft} and Table. \ref{tab:ss_lp}. For both vanilla fine-tuning (Table. \ref{tab:ss_vanilla-ft}) and LFT (Table. \ref{tab:ss_lp}), LDA-based metric demonstrates a 12.7\% and 15.06\% higher average weighted Kendall coefficient compared to the best performing SOTA. The superior performance of the LDA-based metric compared to SOTA metrics across all fine-tuning types emphasizes a key insight: previous metrics have been primarily designed with supervised models in mind, overlooking the unique characteristics and requirements of self-supervised models.  To understand why the LDA-based metric can outperform SOTA on self-supervised estimation tasks, we study the difference in feature geometry generated by the self-supervised and supervised models. The t-SNE feature visualization can be found in Fig. \ref{fig:ss_vs_s}. Without semantic supervision, the self-supervised models present a more heterogeneous feature space than that of the supervised models. This finding also indicates that when estimating the transferability of models with diverse, less discriminative feature space, the existing transferability estimation metric could be vulnerable. On the other hand, the capability of the LDA metric to reflect the transferability of less discriminative feature spaces demonstrates a foundation for developing transferability estimation matrics for self-supervised models.

\section{Conclusions and Discussions}

The introduction and evaluation of our feature perturbation method represent a significant advancement in the field of transferability estimation. Through a comprehensive analysis, we have observed that our feature perturbation method not only enhances the accuracy of existing transferability metrics across various fine-tuning methods but also introduces a vital aspect of robustness evaluation. This additional layer of analysis provides more precise rankings by assessing the resilience of model embeddings to perturbations, ensuring that the best model is robust and transferable to new, target datasets.
Specifically, our method has shown to significantly improve metrics such as $\mathcal{N}$LEEP, LogMe, GBC, SFDA, and NCTI with varied effects on time complexity, indicating its capacity to optimize computational efficiency in certain scenarios.

We provide insights into the disparities in embedding structures between self-supervised and supervised models, emphasizing the need for carefully tailored transferability estimation metrics for both model types. Our results reveal that an LDA-based metric outperforms SOTA across all fine-tuning variants for self-supervised tasks. This highlights an opportunity for the community to develop more adaptable and accurate transferability estimation metrics. 

\section*{Acknowledgment}

This work was supported in part by the Australian Research Council (FT230100426).

%%%%%%%%% REFERENCES
{\small
\bibliographystyle{ieee_fullname}
\bibliography{egbib}
}

\clearpage  % Ensures supplementary material starts on a new page

\appendix
\section*{Supplementary Material}

\input{supplement}

\end{document}

%% file: supplement.tex
The supplementary material contains additional details and further results related to the main paper.

\section{Implementation details of LDA-Based Score}

This section describes the implementation process for deriving a classification score based on Linear Discriminant Analysis (LDA). LDA finds a linear combination of features that best separate the classes of data. The core optimization problem of LDA is expressed as:
\begin{equation}
    U = \text{arg max}_{U} \frac{U^T \Sigma_{\beta} U}{U^T \Sigma_{\omega} U},
\end{equation}
where \( \Sigma_{\beta} \) and \( \Sigma_{\omega} \) represent the between-class and within-class scatter matrices, respectively. This formula aims to project feature vectors to maximize the ratio of between-class variance to within-class variance.

To solve this optimization, we follow the methodology outlined in Ghojogh et al. (2019), which leads to the solution of the generalized eigenvalue problem:

\begin{equation}
    U = \text{eig}((\Sigma_{\omega} + \epsilon I)^{-1} \Sigma_{\beta}),
\end{equation}
where \( \epsilon \) is a small positive scalar that ensures the non-singularity of the within-class scatter matrix \( \Sigma_{\omega} \).

In our transformed feature space, each class's features are assumed to follow a normal distribution centered at their projected class means. Bayes' theorem expresses the score function for a sample for label \( c \) as in Eq. \ref{eq:bayes}. We then compute the LDA-based metric score \( S_{lda} \) using:
\begin{equation}
    \bar{f} = U^T \mathcal{F},
\end{equation}
\begin{equation}
    \delta_c = \bar{f}^T U^T \mu_c - \frac{1}{2} \mu_c^T U U^T \mu_c + \log(\frac{K_c}{K}),
    \label{eq:bayes}
\end{equation}
\begin{equation}
    S_{lda} = \frac{1}{K} \sum_{k=1}^{K} \frac{e^{\delta_y}}{\sum_{c=1}^{C} e^{\delta_c}},
\end{equation}
where \( y \) is the ground-truth class label, \( K \) is the total number of samples, and \( C \) is the number of classes. The \( S_{lda} \) represents the probability of correctly classifying a sample based on the calculated discriminative scores \( \delta_c \) for each class.

\section{More experimental results}

This section presents experimental results that were not included in the main paper due to space limitations.

\subsection{LBFT for self-supervised models}

Besides vanilla fine-tuning and LFT for self-supervised models, LDA-based metrics also demonstrate improved performance in transferability estimation for LBFT. (see Table \ref{tab:ss_LBFT}). The overall average values of these metrics are significantly lower compared to LFT and vanilla fine-tuning. This indicates that the previous metrics are not well-suited for LBFT fine-tuning. Observing Table \ref{tab:lbft_supervised}, it's evident that the Aircraft dataset exhibits the lowest target accuracy among all datasets, suggesting that self-supervised models pre-trained with ImageNet may not be optimally suited for transfer to the Aircraft dataset. This influence is further highlighted in Table \ref{tab:ss_LBFT}, where the performance of all previous metrics for the Aircraft dataset shows negative correlation scores, showing that the metric may not rank the model properly based on the feature embedding.

% \vspace{2em}
\section{Visualization: Correlation between the metric and accuracy}

\begin{figure*}[htbp]
    \centering
    \includegraphics[width=0.95\textwidth]{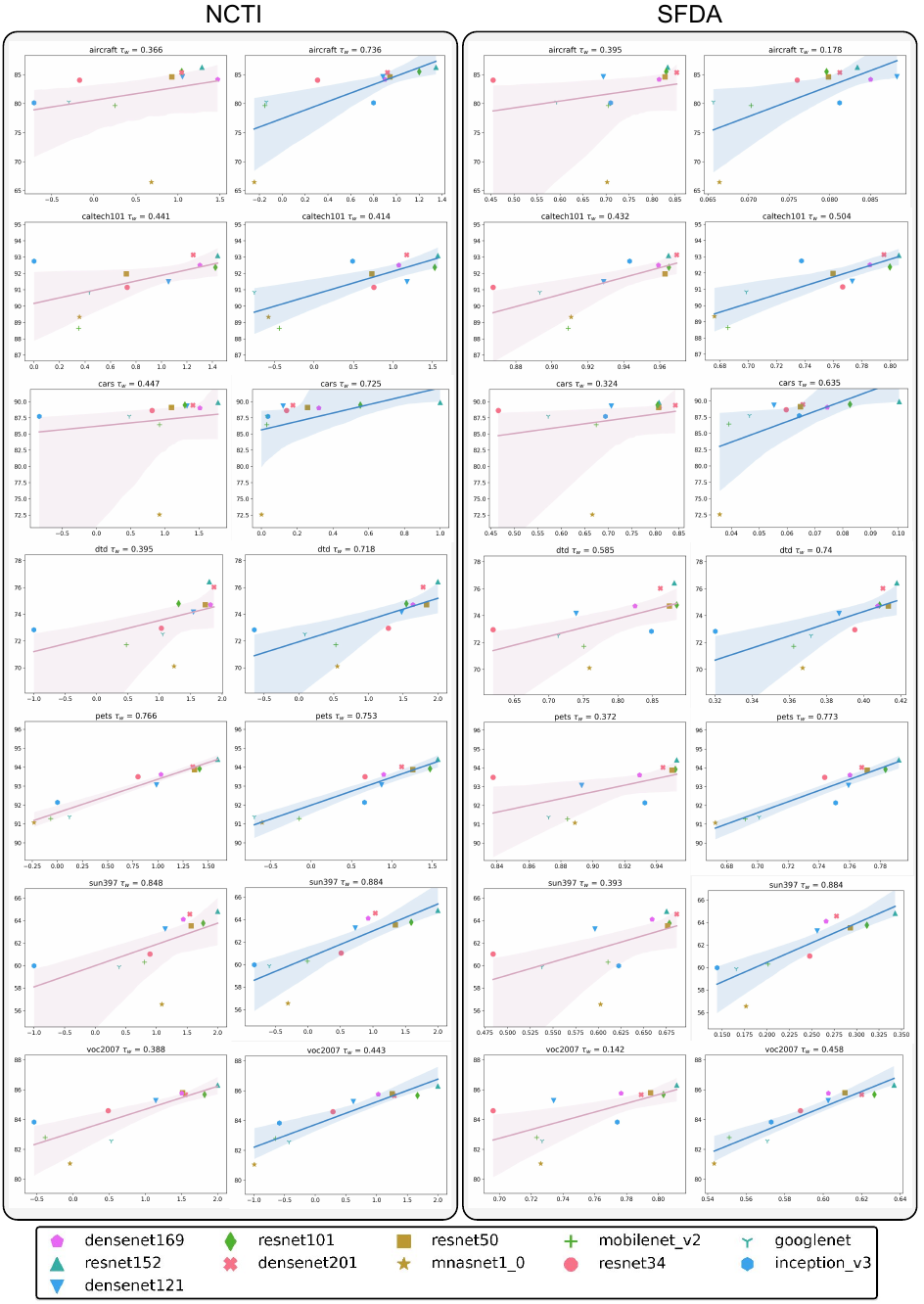}
    \caption{The figure illustrates the correlation between transferability scores and model performance (\%) on the target dataset after vanilla fine-tuning (best viewed in color). Each marker denotes a distinct supervised pre-trained model. We demonstrate an enhancement for NCTI and SFDA using the weighted Kendall $\tau_w$ of our feature perturbation method (in blue) over the original method (in pink).}
    \label{fig:regression plot}
\end{figure*}

Fig. \ref{fig:regression plot} depicts the relationship between the ground truth target accuracy (vanilla fine-tuning) and the transferability estimation metric score across various datasets. We use the best two metrics (\textit{i.e.}, SFDA and NCTI) to illustrate the regression plots for the original metric (depicted in pink) following the application of our feature perturbation method (depicted in blue). The shaded area indicates the 95 confidence interval. After applying our feature perturbation, the width of the shaded region in the regression plot decreases, indicating the metric score and target accuracy are more linearly correlated. Additionally, the ranking of many models shifts, bringing them closer to the shaded area, resulting in an enhancement in the weighted Kendall $\tau_w$. 

% \clearpage

% \begin{figure}[h]
%     \centering
%     \includegraphics[width=0.95\textwidth]{images/Copy of Copy of supplementary_fig.pdf}
%     \caption{The figure illustrates the correlation between transferability scores and model performance (\%) on the target dataset after vanilla fine-tuning. Each marker denotes a distinct supervised pre-trained model. We demonstrate an enhancement for NCTI and SFDA using the weighted Kendall $\tau_w$ of our feature refinement method (in blue) over original method (in pink).} 
%     \label{fig:regression plot}
% \end{figure}

\section{Downstream datasets description}
We validate the effectiveness of the proposed methods on 11 standard datasets commonly adopted in transferability estimation metric evaluation. The datasets can be categorized as follows:

\begin{enumerate}
    \item \textbf{Fine-grained classification datasets:}
    \begin{itemize}
    \vspace{1em}
        \item FGVC Aircraft: This dataset contains images of various aircraft types for fine-grained classification. It consists of 100 classes with a total of 10,000 images, split into a 2:3 ratio for training and testing.
        \item Stanford Cars: Comprising images of cars from different viewpoints, this dataset totals 16,185 images across various car brands and models, providing a diverse set of images for training and evaluation.
        The training set contains 8,144 images, while the test set contains 8,041 images.
        \item Food-101: A dataset with 101,000 images categorized into 101 food classes. Each food category contains 750 training images and 250 testing images.
        \item Oxford-IIIT Pets: This dataset includes 7,049 pet images belonging to 37 different pet breeds with a varying number of images per breed. The training set consists of 3,680 images, and the testing set has 3,669 images.
        \item Oxford-102 Flowers: It has 102 categories with varying numbers of images per category. It comprises between 40 and 258 images per category, with 20 images sampled for training and the remaining 6,149 images for testing.
    \end{itemize}
    
    \item \textbf{Coarse-grained classification dataset:}
    \begin{itemize}
    \vspace{1em}
        \item Caltech-101: A dataset with 9,146 images distributed among 101 categories. 70\% of the data is sampled for the training set.
        \item CIFAR-10 and CIFAR-100: These datasets contain 60,000 color images of object categories, including animals, vehicles, and everyday objects, making them suitable for general-purpose image classification. CIFAR-10 is divided into 10 distinct classes with 5,000 training images and 1,000 testing images per class. CIFAR-100 is divided into 100 distinct classes with 500 training images and 100 testing images per class.
        \item VOC2007: This dataset consists of 9,963 images across 20 classes with a variety of common object classes, including people, animals, vehicles, and household items. The training set comprises 5,011 images and the remaining for testing.
    \end{itemize}
    
    \item \textbf{Scene classification dataset:}
    \begin{itemize}
    \vspace{1em}
        \item SUN397: This dataset contains 397 classes, each with 1,000 scenery pictures, totaling 19,850 images. The dataset covers a wide range of scenes, including indoor/outdoor environments, and natural/urban settings.
    \end{itemize}
    
    \item \textbf{Texture classification dataset:}
    \begin{itemize}
    \vspace{1em}
        \item DTD: This dataset includes 5,640 textural images categorized into 47 classes. The dataset includes high-quality images with variations in lighting, scale, and orientation, making it suitable for studying the challenges of texture recognition in real-world scenarios. Each class contains 80 training images and 40 testing images.
    \end{itemize}
\end{enumerate}

\section{Ground Truth}

The ground truth target accuracies of vanilla FT, LBFT, and LFT for supervised models are given in Table. \ref{tab:vanilla_ft_supervised}, \ref{tab:lbft_supervised}, \ref{tab:lft_supervised}. For self-supervised models, the ground truth target accuracies are given in Table. \ref{tab:ground_truth-vanilla_ft}, \ref{tab:ground_truth-LBFT}, \ref{tab:ground_truth-LFT}. For both LBFT and LFT, we follow the grid search described in [34], which selects the learning rates from $\{10^{-1}, 10^{-2}, 10^{-3}, 10^{-4}\}$ and weight decay parameters from $\{10^{-6}, 10^{-5}, 10^{-4}, 10^{-3}\}$. Once the optimal hyper-parameters are identified, we proceed to fine-tune the pre-trained model on the designated dataset using these hyper-parameters. The resulting test accuracy serves as our benchmark. Fine-tuning is conducted on a NVIDIA A100, utilizing a batch size of 128, and all input images are resized to dimensions of 224×224.  To ensure the robustness and reliability of our evaluation, we execute the code using five distinct seeds for each experiment.

 % self sup conv
\begin{table*}
    \setlength\tabcolsep{5pt}
    \footnotesize
    \centering
    \caption{Performance comparison (average weighted Kendall $\tau_w$) for LBFT on self-supervised models. 
    The highest performing $\tau_w$ value in each column are highlighted in bold. LDA achieves the highest overall average weighted Kendall $\tau_w$ score.}
    \label{tab:ss_LBFT}
    \begin{adjustbox}{max width=\textwidth}
    \begin{tabular}{l|ccccccccccc|c}
        \toprule
        Method & Aircraft & Caltech-101 & Cars & CIFAR10 & CIFAR100 & DTD & Flowers & Food-101 & Pets & Sun397 & VOC & Average \\
        \midrule
        $\mathcal{N}$LEEP [24] & -0.288 & 0.600 & 0.277 & \textbf{0.291} & 0.326 & 0.849 & 0.283 & 0.603 & 0.683 & 0.099 & 0.184 & 0.355 \\
        LogME [43]  & \textbf{-0.117} & 0.326 & 0.237 & -0.177 & -0.183 & 0.836 & \textbf{0.512} & \textbf{0.645} & 0.708 & -0.157 & 0.407 & 0.276 \\
        GBC [29]     & -0.244 & 0.342 & 0.171 & 0.263 & 0.264 & 0.472 & 0.417 & 0.495 & 0.485 & 0.247 & 0.408 & 0.302 \\
        SFDA [33]    & -0.189 & 0.465 & 0.088 & -0.056 & 0.056 & 0.707 & 0.357 & 0.550 & 0.738 & 0.163 & 0.688 & 0.324 \\
        NCTI [39]    & -0.194 & 0.610 & 0.046 & -0.077 & \textbf{0.428} &\textbf{0.895} & 0.372 & 0.421 & \textbf{0.757} & 0.256 & \textbf{0.738} & 0.387 \\
        LDA & -0.282 & \textbf{0.655} & \textbf{0.720} & 0.169 & 0.330 & 0.686 & 0.353 & 0.462 & 0.670 & \textbf{0.588} & 0.681 & \textbf{0.389} \\
        \bottomrule
    \end{tabular}
    \end{adjustbox}

\end{table*}

\begin{table*}\setlength\tabcolsep{5pt}
\footnotesize
    \centering
    \caption{The ground truth target accuracy of vanilla fine-tuning for supervised models on 11 target datasets is sourced from [33].}
    \label{tab:vanilla_ft_supervised}
    \begin{adjustbox}{max width=\textwidth}
    \begin{tabular}{lcccccccccccc}
        \toprule
        & Aircraft & Caltech-101 & Cars  & CIFAR10 & CIFAR100 & DTD   & Flowers & Food-101 & Pets  & Sun & VOC  \\
        \midrule
        ResNet-34 & 84.06 & 91.15 & 88.63 & 96.12 & 81.94 & 72.96 & 95.2 & 81.99 & 93.5 & 61.02 & 84.6 \\
        ResNet-50 & 84.64 & 91.98 & 89.09 & 96.28 & 82.8 & 74.72 & 96.26 & 84.45 & 93.88 & 63.54 & 85.8 \\
        ResNet-101 & 85.53 & 92.38 & 89.47 & 97.39 & 84.88 & 74.8 & 96.53 & 85.58 & 93.92 & 63.76 & 85.68 \\
        ResNet-152 & 86.29 & 93.1 & 89.88 & 97.53 & 85.66 & 76.44 & 96.86 & 86.28 & 94.42 & 64.82 & 86.32 \\
        DenseNet-121 & 84.66 & 91.5 & 89.34 & 96.45 & 82.75 & 74.18 & 97.02 & 84.99 & 93.07 & 63.26 & 85.28 \\
        DenseNet-169 & 84.19 & 92.51 & 89.02 & 96.77 & 84.26 & 74.72 & 97.32 & 85.84 & 93.62 & 64.1 & 85.77 \\
        DenseNet-201 & 85.38 & 93.14 & 89.44 & 97.02 & 84.88 & 76.04 & 97.1 & 86.71 & 94.03 & 64.57 & 85.67 \\
        MNet-A1 & 66.48 & 89.34 & 72.58 & 92.59 & 72.04 & 70.12 & 95.39 & 71.35 & 91.08 & 56.56 & 81.06 \\
        MobileNetV2 & 79.68 & 88.64 & 86.44 & 94.74 & 78.11 & 71.72 & 96.2 & 81.12 & 91.28 & 60.29 & 82.8 \\
        Googlenet & 80.32 & 90.85 & 87.76 & 95.54 & 79.84 & 72.53 & 95.76 & 79.3 & 91.38 & 59.89 & 82.58 \\
        InceptionV3 & 80.15 & 92.75 & 87.74 & 96.18 & 81.49 & 72.85 & 95.73 & 81.76 & 92.14 & 59.98 & 83.84 \\
        \bottomrule
    \end{tabular}
    \end{adjustbox}
\end{table*}

\begin{table*}\setlength\tabcolsep{5pt}
\footnotesize
    \centering
    \caption{The ground truth target accuracy of LBFT for supervised models on 11 target datasets.}
    \label{tab:lbft_supervised}
    \begin{adjustbox}{max width=\textwidth}
    \begin{tabular}{lcccccccccccc}
        \toprule
        & Aircraft & Caltech-101 & Cars  & CIFAR10 & CIFAR100 & DTD   & Flowers & Food-101 & Pets  & Sun & VOC  \\
        \midrule
        InceptionV3 & 47.98 & 90.25 & 56.6 & 83.76 & 63.46 & 68.99 & 90.76 & 63.77 & 87.78 & 81.6 & 80.88 \\
        MobileNetV2 & 53.33 & 86.78 & 69.83 & 87.06 & 66.59 & 73.19 & 94.95 & 72.24 & 90.41 & 83.26 & 82.03 \\
        MNet-A1 & 52.05 & 88.92 & 65.08 & 74.4 & 40.68 & 68.03 & 93.95 & 67.05 & 90.54 & 73 & 82.37 \\
        DenseNet-121 & 67.81 & 90.24 & 81.72 & 93.7 & 78.23 & 72.93 & 97.36 & 79.29 & 91.26 & 90.39 & 84.42 \\
        DenseNet-169 & 74.29 & 92.51 & 83.94 & 96.06 & 84.2 & 74.36 & 96.47 & 82.06 & 93.65 & 96.83 & 86.03 \\
        DenseNet-201 & 71.51 & 92.02 & 83.51 & 95.74 & 83.85 & 74.41 & 97.36 & 81.94 & 91.89 & 97.02 & 85.36 \\
        ResNet-34 & 70.94 & 90.42 & 83.07 & 93.94 & 80.74 & 71.54 & 96.42 & 78.04 & 92.84 & 94.84 & 84.39 \\
        ResNet-50 & 76.43 & 91.4 & 84.93 & 86.49 & 84.46 & 74.57 & 97.25 & 82.8 & 93.87 & 96.28 & 85.67 \\
        ResNet-101 & 75.57 & 91.92 & 85.19 & 96.34 & 85.04 & 74.79 & 96.48 & 83.02 & 93.44 & 97.41 & 85.76 \\
        ResNet-152 & 74.68 & 92.45 & 85.75 & 96.18 & 84.73 & 75.16 & 95.41 & 82.86 & 93.93 & 96.56 & 86.15 \\
        Googlenet & 64.55 & 90.31 & 78.06 & 92.67 & 76.1 & 72.82 & 95.08 & 72.69 & 89.67 & 92.4 & 80.75 \\
        \bottomrule
    \end{tabular}
    \end{adjustbox}
\end{table*}

\begin{table*}\setlength\tabcolsep{5pt}
\footnotesize
\centering
\caption{The ground truth target accuracy of LFT for supervised models on 11 target datasets.}
\label{tab:lft_supervised}
\begin{adjustbox}{max width=\textwidth}
\begin{tabular}{lcccccccccccc}
\toprule
& Aircraft & Caltech-101 & Cars & CIFAR10 & CIFAR100 & DTD & Flowers & Food-101 & Pets & Sun & VOC \\
\midrule
InceptionV3 & 28.21 & 88.48 & 27.6 & 69.87 & 46.39 & 61.28 & 83.01 & 46.31 & 85.85 & 63.72 & 77.01 \\
MobileNetV2 & 42.24 & 87.35 & 49.77 & 76.97 & 57.46 & 67.77 & 92.27 & 62.6 & 89.73 & 73.25 & 80.88 \\
MNet-A1 & 41.72 & 87.85 & 46.19 & 69.55 & 37.49 & 65.69 & 92.37 & 62.65 & 89.56 & 79.63 & 81.18 \\
DenseNet-121 & 43.61 & 90.03 & 51.78 & 81.39 & 62.11 & 68.09 & 93.23 & 65.37 & 91.46 & 76.37 & 82.73 \\
DenseNet-169 & 47.15 & 90.76 & 56.2 & 83.08 & 64.53 & 69.95 & 94.15 & 67.81 & 92.6 & 80.78 & 84.07 \\
DenseNet-201 & 46.39 & 91.31 & 57.32 & 84.52 & 67.51 & 70.64 & 93.01 & 68.11 & 92.57 & 80.38 & 83.34 \\
ResNet-34 & 38.19 & 89.8 & 32.04 & 78.61 & 59.43 & 66.7 & 90.71 & 60.56 & 91.27 & 71.96 & 82.46 \\
ResNet-50 & 40.63 & 89.75 & 50.91 & 83.57 & 65.41 & 70.74 & 93.05 & 65.79 & 91.76 & 83.29 & 83.28 \\
ResNet-101 & 41.21 & 89.81 & 50.6 & 85.24 & 67.64 & 69.57 & 92.3 & 66.5 & 92.34 & 75.61 & 83.85 \\
ResNet-152 & 42.98 & 91.42 & 52.07 & 85.33 & 67.81 & 70.74 & 93.06 & 67.55 & 92.67 & 75.72 & 84.13 \\
Googlenet & 36.22 & 88.31 & 43.83 & 78.45 & 59.73 & 66.12 & 89.53 & 55.34 & 89.41 & 76.82 & 80.32 \\
\bottomrule
\end{tabular}
\end{adjustbox}
\end{table*}

\begin{table*}\setlength\tabcolsep{5pt}
\footnotesize

    \centering
    \caption{The ground truth target accuracy of vanilla fine-tuning for self-supervised models on 11 target datasets is sourced from [33].}
    \label{tab:ground_truth-vanilla_ft}
    \begin{adjustbox}{max width=\textwidth}

    \begin{tabular}{lcccccccccccc}
        \toprule
        & Aircraft & Caltech-101 & Cars  & CIFAR10 & CIFAR100 & DTD   & Flowers & Food-101 & Pets  & Sun & VOC  \\
        \midrule
        BYOL & 82.10 & 91.90 & 89.83 & 96.98 & 83.86 & 76.37 & 96.80 & 85.44 & 91.48 & 63.69 & 85.13 \\
        Deepclusterv2 & 82.43 & 91.16 & 90.16 & 97.17 & 84.84 & 77.31 & 97.05 & 87.24 & 90.89 & 66.54 & 85.38 \\
        Infomin & 83.78 & 80.86 & 86.90 & 96.72 & 70.89 & 73.47 & 95.81 & 78.82 & 90.92 & 57.67 & 81.41 \\
        InsDis & 79.70 & 77.21 & 80.21 & 93.08 & 69.08 & 66.40 & 93.63 & 76.47 & 84.58 & 51.62 & 76.33 \\
        MoCov1 & 81.85 & 79.68 & 82.19 & 94.15 & 71.23 & 67.36 & 94.32 & 77.21 & 85.26 & 53.83 & 77.94 \\
        MoCov2 & 83.70 & 82.76 & 85.55 & 96.48 & 71.27 & 72.56 & 95.12 & 77.15 & 89.06 & 56.28 & 78.32 \\
        PCLv1 & 82.16 & 88.60 & 87.15 & 96.42 & 79.44 & 73.28 & 95.62 & 77.70 & 88.93 & 58.36 & 81.91 \\
        PCLv2 & 83.00 & 87.52 & 85.56 & 96.55 & 79.84 & 69.3 & 95.87 & 80.29 & 88.72 & 58.82 & 81.85 \\
        Sela-v2 & 85.42 & 90.53 & 89.85 & 96.85 & 84.36 & 76.03 & 96.22 & 86.37 & 89.61 & 65.74 & 85.52 \\
        SimCLRv1 & 80.54 & 90.94 & 89.98 & 97.09 & 84.49 & 73.97 & 95.33 & 82.2 & 88.53 & 63.46 & 83.29 \\
        SimCLRv2 & 81.50 & 88.58 & 88.82 & 96.22 & 78.91 & 74.71 & 95.39 & 82.23 & 89.18 & 60.93 & 83.08 \\
        SWAV & 83.04 & 89.49 & 89.81 & 96.81 & 83.78 & 76.68 & 97.11 & 87.22 & 90.59 & 66.10 & 85.06 \\
        \bottomrule
    \end{tabular}
    \end{adjustbox}
\end{table*}

\begin{table*}\setlength\tabcolsep{5pt}
\footnotesize

    \centering
    \caption{The ground truth target accuracy of LBFT for self-supervised models on 11 target datasets.}
    \label{tab:ground_truth-LBFT}
    \begin{adjustbox}{max width=\textwidth, margin=0em}
    \begin{tabular}{lcccccccccccc}
        \toprule
        & Aircraft & Caltech-101 & Cars  & CIFAR10 & CIFAR100 & DTD   & Flowers & Food-101 & Pets  & Sun & VOC  \\
        \midrule
        BYOL & 73.09 & 91.05 & 85.15 & 98.50 & 92.52 & 74.68 & 96.16 & 83.11 & 89.60 & 99.81 & 84.06 \\
        Deepclusterv2 & 71.16 & 89.62 & 83.26 & 95.95 & 84.62 & 75.21 & 95.87 & 83.29 & 89.82 & 99.79 & 84.59 \\
        Infomin & 77.46 & 84.77 & 85.57 & 96.31 & 82.15 & 74.63 & 96.18 & 84.25 & 88.60 & 98.62 & 82.56 \\
        InsDis & 70.65 & 76.14 & 79.48 & 94.16 & 77.20 & 70.74 & 91.91 & 79.05 & 80.28 & 97.96 & 76.50 \\
        MoCov1 & 73.16 & 78.40 & 81.40 & 94.35 & 77.97 & 71.49 & 92.24 & 78.70 & 83.05 & 98.02 & 78.10 \\
        MoCov2 & 75.70 & 85.18 & 84.37 & 96.23 & 82.12 & 73.46 & 95.47 & 82.57 & 87.78 & 97.87 & 81.20 \\
        PCLv1 & 76.60 & 85.63 & 83.82 & 96.94 & 85.98 & 73.03 & 94.78 & 80.82 & 85.83 & 99.13 & 80.87 \\
        PCLv2 & 76.65 & 85.07 & 84.94 & 97.75 & 87.97 & 72.45 & 95.11 & 82.62 & 87.05 & 99.17 & 81.42 \\
        Sela-v2 & 69.85 & 87.56 & 81.66 & 95.56 & 83.75 & 74.36 & 94.94 & 82.62 & 88.54 & 99.53 & 85.19 \\
        SimCLRv1 & 67.45 & 90.48 & 77.08 & 96.96 & 87.65 & 71.65 & 92.13 & 75.77 & 85.43 & 99.48 & 82.29 \\
        SimCLRv2 & 74.04 & 85.19 & 84.83 & 97.38 & 89.90 & 72.45 & 95.50 & 83.09 & 84.92 & 99.85 & 80.76 \\
        SWAV & 71.65 & 88.49 & 82.84 & 95.72 & 83.69 & 75.85 & 95.66 & 83.31 & 87.62 & 99.80 & 84.22 \\
        \bottomrule
    \end{tabular}
\end{adjustbox}
\end{table*}

\begin{table*}\setlength\tabcolsep{5pt}
\footnotesize
    \centering
    \caption{The ground truth target accuracy of LFT for self-supervised models on 11 target datasets.}
    \label{tab:ground_truth-LFT}
    \begin{adjustbox}{max width=\textwidth, margin=0em}
    \begin{tabular}{lcccccccccccc}
        \toprule
        & Aircraft & Caltech-101 & Cars  & CIFAR10 & CIFAR100 & DTD   & Flowers & Food-101 & Pets  & Sun & VOC  \\
        \midrule
        BYOL & 43.48 & 89.83 & 43.45 & 84.07 & 57.71 & 71.28 & 92.75 & 61.17 & 87.13 & 66.36 & 74.79 \\
        Deepclusterv2 & 47.44 & 89.34 & 56.19 & 79.43 & 55.19 & 72.45 & 93.65 & 68.62 & 87.08 & 81.41 & 80.91 \\
        Infomin & 12.81 & 80.61 & 7.24 & 58.89 & 22.09 & 65.11 & 63.58 & 37.98 & 80.96 & 38.14 & 74.28 \\
        InsDis & 10.93 & 51.26 & 3.82 & 42.81 & 15.65 & 56.33 & 58 & 27.06 & 50.77 & 31.08 & 52.17 \\
        MoCov1 & 10.88 & 54.23 & 3.32 & 45.01 & 15.68 & 54.41 & 54.56 & 26.89 & 53.03 & 31.01 & 55.92 \\
        MoCov2 & 11.51 & 78.43 & 5.52 & 54.22 & 24.09 & 64.89 & 59.73 & 34.86 & 73.62 & 34.81 & 70.54 \\
        PCLv1 & 7.46 & 70.13 & 3.90 & 50.70 & 22.68 & 52.23 & 36.81 & 21.12 & 68.08 & 25.84 & 67.99 \\
        PCLv2 & 13.99 & 82.41 & 8.20 & 69.79 & 32.66 & 65.90 & 69.71 & 36.15 & 75.51 & 39.06 & 72.15 \\
        Sela-v2 & 31.31 & 84.62 & 24.40 & 73.00 & 38.91 & 72.07 & 87.64 & 58.06 & 82.27 & 65.42 & 77.46 \\
        SimCLRv1 & 42.75 & 88.72 & 43.23 & 83.77 & 61.60 & 67.07 & 88.42 & 58.55 & 79.86 & 82.51 & 78.87 \\
        SimCLRv2 & 39.96 & 86.66 & 42.54 & 80.74 & 55.51 & 71.97 & 91.34 & 63.24 & 81.79 & 76.51 & 77.76 \\
        SWAV & 43.25 & 87.85 & 45.94 & 75.93 & 47.59 & 74.15 & 92.54 & 66.42 & 85.23 & 77.48 & 79.28 \\
        \bottomrule
    \end{tabular}
\end{adjustbox}

\end{table*}

